\documentclass[10pt,twocolumn,letterpaper]{article}
\usepackage[utf8]{inputenc}
\usepackage{cvpr}
\usepackage{times}
\usepackage{epsfig}
\usepackage{graphicx}
\usepackage{amsmath}
\usepackage{amssymb}

\usepackage[backend=bibtex, style=ieee]{biblatex}
\usepackage{hyperref}
\hypersetup{
    bookmarks=true,         
    unicode=false,          
    pdftoolbar=true,        
    pdfmenubar=true,        
    pdffitwindow=false,     
    pdfstartview={FitH},    
    pdftitle={Exploring the Neural Algorithm of Artistic Style},    
    pdfauthor={Yaroslav Nikulin and Roman Novak},     
    pdfsubject={Object Recognition and Computer Vision},   
    pdfkeywords={Style Transfer, Deep Convolutional Neural Networks, Image Processing}, 
    pdfnewwindow=true,      
    colorlinks=true,       
    linkcolor=red,          
    citecolor=green,        
    filecolor=magenta,      
    urlcolor=cyan           
}

\cvprfinalcopy 
\def\cvprPaperID{****} 

\begin{document}

\title{Exploring the Neural Algorithm of Artistic Style}
\author{
\textbf{Yaroslav Nikulin}\thanks{Authors contributed equally to this work. }  \textnormal{ and} \textbf{Roman Novak}$^*$\\
Department of Mathematics\\
École normale supérieure de Cachan\\
94230 Cachan, France \\
{\texttt{\{}\href{mailto:yaroslav.nikulin@ens-cachan.fr}{\texttt{yaroslav.nikulin}}\texttt{, }\href{mailto:rnovak@ens-cachan.fr}{\texttt{rnovak}}\}\texttt{@ens-cachan.fr}}
}

\maketitle

\begin{abstract}
   In this work we explore the method of style transfer presented in \cite{Gatys15}. 
   We first demonstrate the power of the suggested style space on a few examples.
   
   We then vary different hyper-parameters and program properties that were not discussed in \cite{Gatys15}, among which are the recognition network used, starting point of the gradient descent and different ways to partition style and content layers. We also give a brief comparison of some of the existing algorithm implementations and deep learning frameworks used.
   
	To study the style space further, an idea similar to \cite{DD} is used to generate synthetic images by maximizing
   a single entry in one of the Gram matrices $\mathcal{G}_l$ and some interesting results are observed. Next, we try to
   mimic the sparsity and intensity distribution of Gram matrices obtained from a real painting
   and generate more complex textures.
   
   Finally, we propose two new style representations 
   built on top of network's features and discuss how one could be used to achieve local and potentially content-aware style transfer.
\end{abstract}

\section{Introduction}
	The key idea behind style transfer is to perform a gradient descent from random noise minimizing the deviation from content (i.e. feature responses in the upper convolutional layers of a recognition network) of the target image and the deviation from the style representation of the style image. The latter is defined as a set of Gram correlation matrices $\left\{\mathcal{G}_l\right\}$ with the entries $\mathcal{G}^l_{i j} = \langle F^l_i, F^l_j\rangle$, where $l$ is the network layer, $i$ and $j$ are two filters of the layer $l$ and $F^l_k$ is the array (indexed by  spatial coordinates) of neuron responses of filter $k$ in layer $l$. In other words, $\mathcal{G}^l_{i j}$ value says how often features $i$ and $j$ in the layer $l$ appear together. Please refer to \cite{Gatys15} for details.

\begin{figure*}
		\begin{center}
			\includegraphics[width=0.33\textwidth]{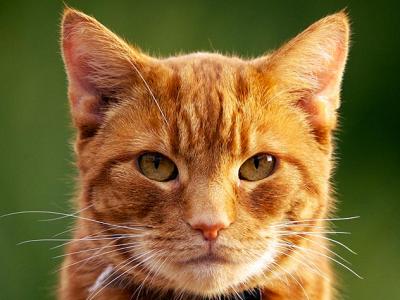}
			\includegraphics[width=0.33\textwidth]{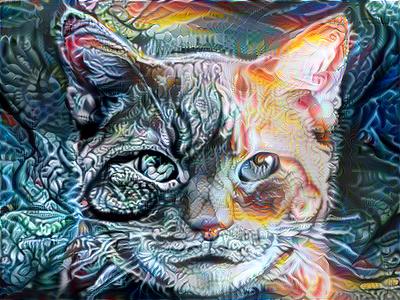}
			\includegraphics[width=0.33\textwidth]{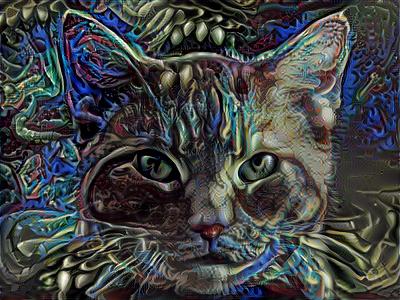}
			\includegraphics[width=0.33\textwidth]{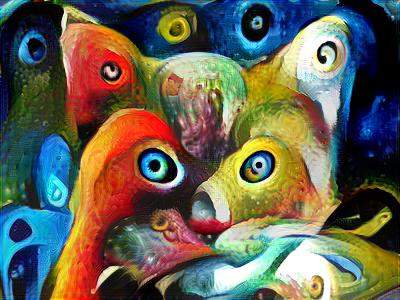}
			\includegraphics[width=0.33\textwidth]{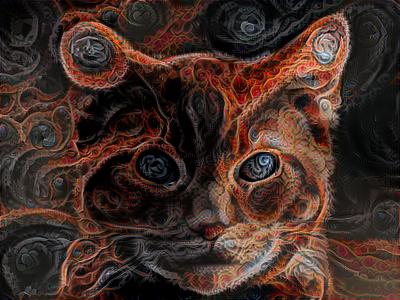}
			\includegraphics[width=0.33\textwidth]{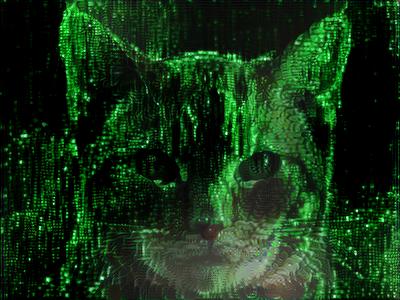}
			\includegraphics[width=0.33\textwidth]{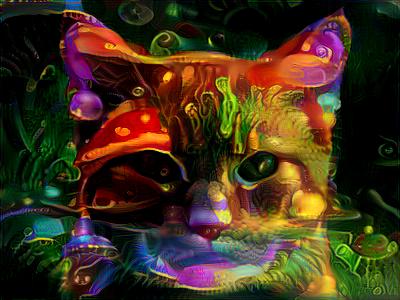}
			\includegraphics[width=0.33\textwidth]{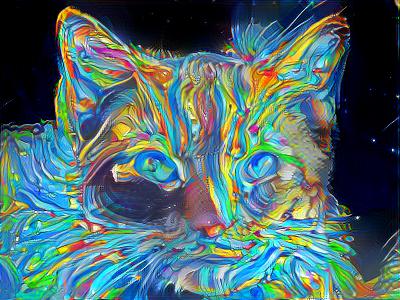}
			\includegraphics[width=0.33\textwidth]{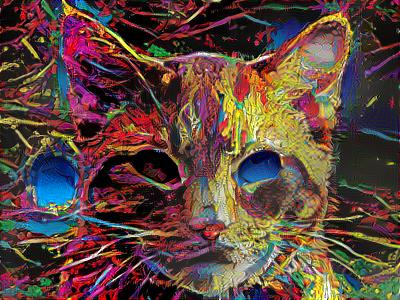}
			\includegraphics[width=0.33\textwidth]{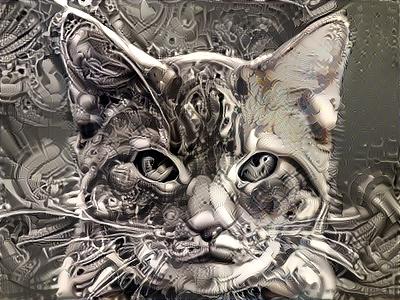}
			\includegraphics[width=0.33\textwidth]{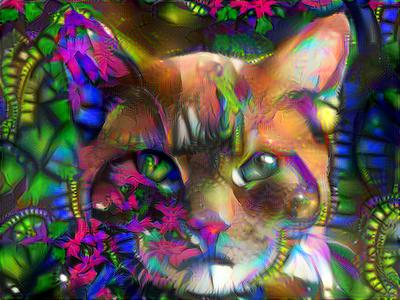}
			\includegraphics[width=0.33\textwidth]{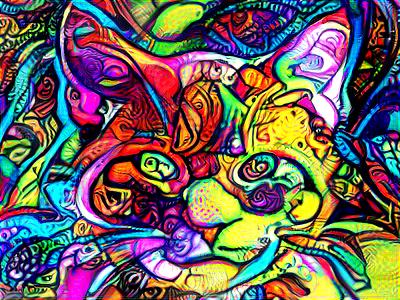}
		\end{center}
		\caption{Transferring different styles \cite{s1, s2, s3, s4, s5, s6, s7, s8, s9, s10, s11} to a photo of a cat  \cite{cat} (top-left).}
		\label{fig:cat}
	\end{figure*}

	On figure \ref{fig:cat} we demonstrate the spectacular performance of the algorithm by running $500$ iterations of the L-BFGS \cite{lbfgs} optimization algorithm on a photo of a cat.
	
	In section \ref{sec:ill} we consider cases where the algorithm doesn't work and suggest possible solutions in section \ref{sec:content}.

\section{Frameworks and Implementations}
	We have tried running implementations on Caffe \cite{Caffe} (using CUDA and OpenCL \cite{CaffeOpenCL} backend) by Frank Liu \cite{fliu} and on Torch \cite{Torch} by Kai Sheng Tai \cite{tai} and by Justin Johnson \cite{jc}.
	
	We have observed the OpenCL Caffe backend to be very promising but yet unstable compared to CUDA. Otherwise, both Torch implementations turned out to be significantly faster.
	
	We have thus built our work on top of the Torch implementation by Kai Sheng Tai. Interestingly, the Torch cunn \cite{cunn} backend performed slightly better than cuDNN \cite{cudnn} by NVIDIA.

\section{Networks}
\begin{figure*}
		\begin{center}
			\includegraphics[width=0.49\textwidth]{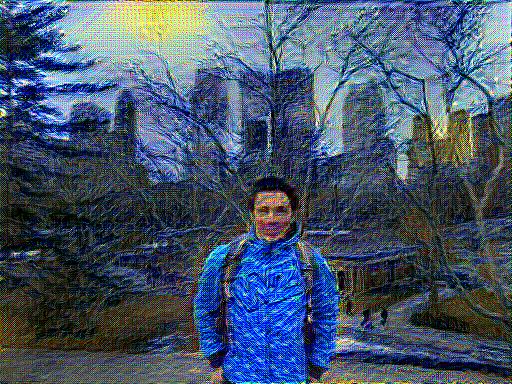}
			\includegraphics[width=0.49\textwidth]{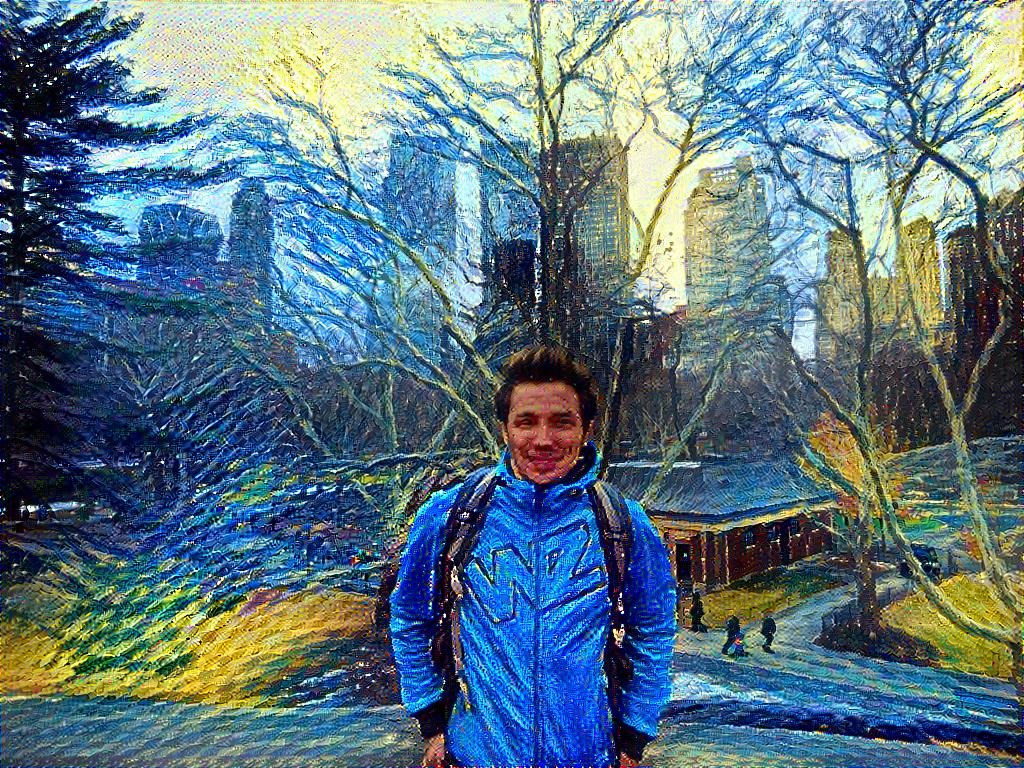}
		\end{center}
		\caption{Style transfer using GoogLeNet (left) and VGG-19 (right).}
		\label{fig:comp}
	\end{figure*}

	In \cite{Gatys15} the VGG-19 \cite{VGG} recognition network is used to produce results. We compare the impact of using other networks (AlexNet \cite{AlexNet}, GoogLeNet \cite{GoogLeNet}, VGG-16 and VGG-19) in figure \ref{fig:comp}, with AlexNet performing similarly to GoogLeNet and VGG-16 similarly to VGG-19.
	
	VGG networks perform much better at style transfer due to their architecture. For example, AlexNet and GoogLeNet strongly compress the input at the  first convolutional layer using large kernels and stride ($11\times 11$  with stride $4$ and $7 \times 7$ with stride 2 respectively) and thus a lot of fine detail is lost. VGG networks use $3\times 3$ kernels with stride 1 at all convolutional layers and thus capture much more information.
	
	We have therefore used the VGG-19 network for all our experiments.

\section{Initialization}
\begin{figure*}
		\begin{center}
			\includegraphics[width=0.495\textwidth]{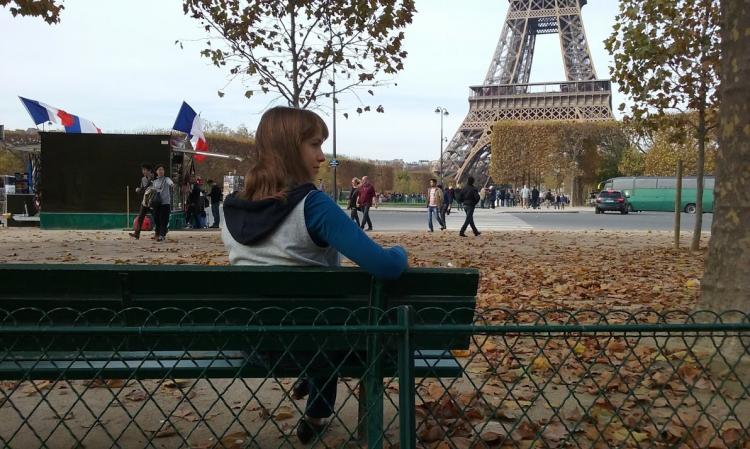}
			\includegraphics[width=0.495\textwidth]{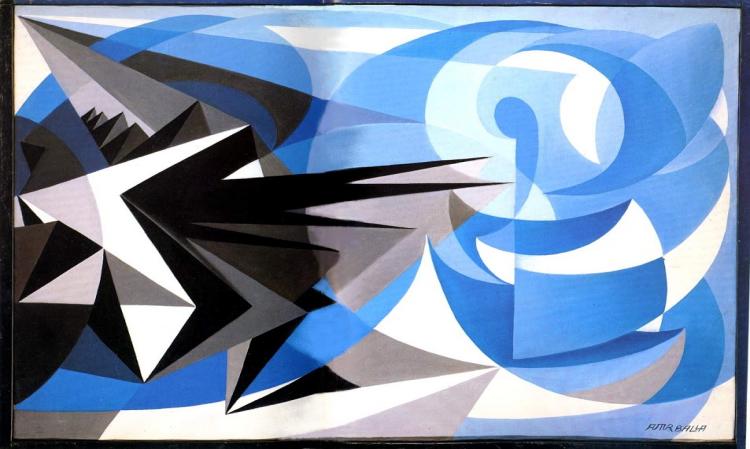}
			\includegraphics[width=0.33\textwidth]{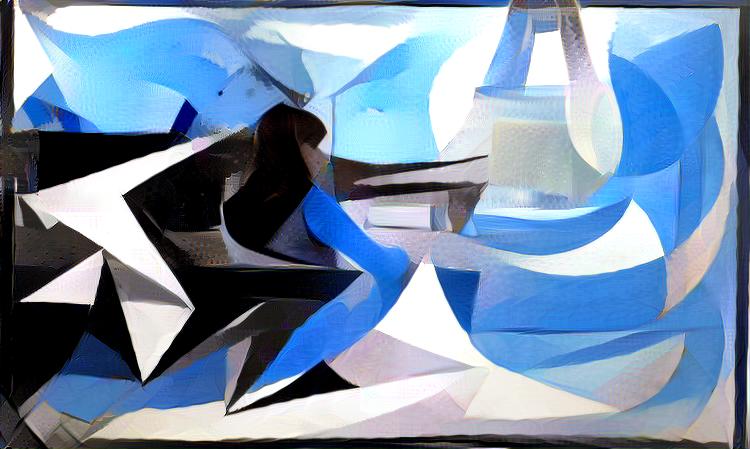}
			\includegraphics[width=0.33\textwidth]{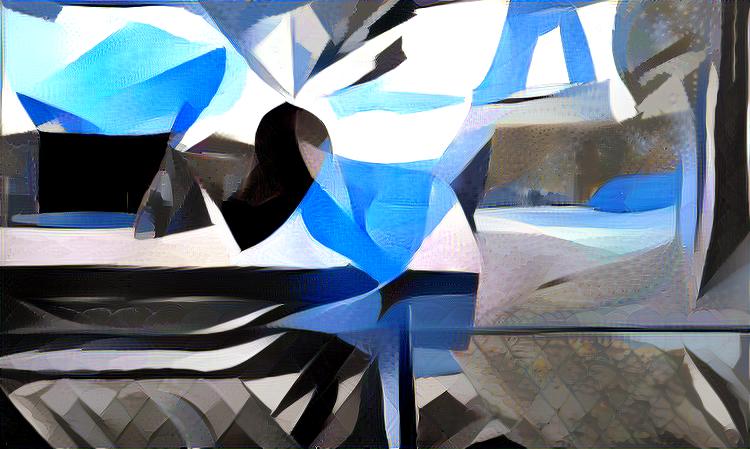}
			\includegraphics[width=0.33\textwidth]{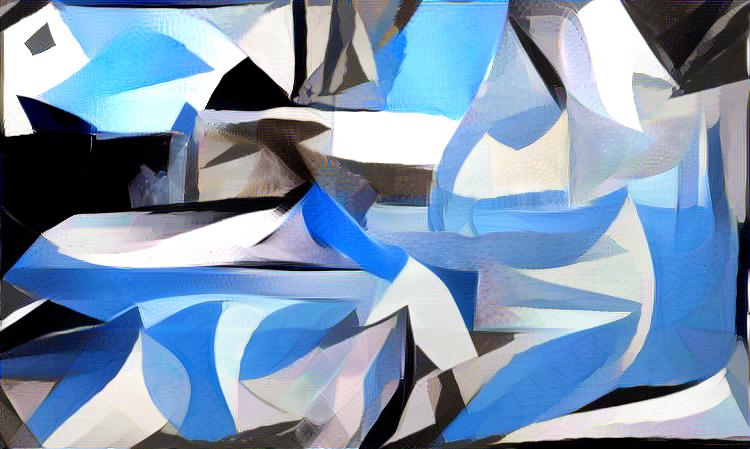}
		\end{center}
		\caption{Impact of the initialization point on the final result. Top row: the source photo (left), the style image (right, \cite{opt}). Bottom row: results of 500 iterations starting from style (left), content (center) and noise (right) images.}
		\label{fig:init}
	\end{figure*}

	In \cite{Gatys15} gradient descent is always performed from white noise. We try and compare two other initialization points: content and style. We demonstrate the impact of the initialization strategy in figure \ref{fig:init}, starting the gradient descent from the content photo, the style artwork or from white noise.
	
	The results highlight well the highly non-convex nature of our objective and that the starting point has a tremendous influence on the basin of attraction that the gradient descent will converge to. We naturally observe that starting from the content lets us preserve the most of it, while starting from the style image is prone to ``leaking'' the content of the artwork into the final result. This also serves to reinforce the observation made in \cite{Gatys15} that style and content are not strictly separable.
	
	We find that for most practical applications starting from the content image produces the best result. This corresponds well to the way most artworks are produced -- starting from the artist observing the content and drawing a rough sketch (i.e. content reconstruction comes first), and only then applying paint (i.e. style) on top.
	
	Note that noise initialization is still very useful for testing, benchmarking and hyper-parameter tuning. For example, starting from content and observing no change one might wonder whether the content weight is too high or the learning rate is too small. Such questions do not occur when starting from noise.

\section{Partial Style Transfer}
	In \cite{Gatys15} the impact of shrinking the style layer pool (from using the first 5 convolutional layers $\{\texttt{conv1-1,...,conv5-1}\}$ down to using only the first convolutional layer $\{\texttt{conv1-1}\}$) is demonstrated. The authors observe how the style feature scale and complexity goes down as they consider using lower and lower convolutional layers to represent style. In general, one doesn't benefit from using only the lower layer style features (apart from the computation facilitation) and is better off simply reducing the style weight in the optimization objective.
	
	In our work we consider keeping the upper convolutional layers and removing the bottom ones, while enforcing them as part of the content pool.
	
\begin{figure*}
		\begin{center}
			\includegraphics[width=0.49\textwidth]{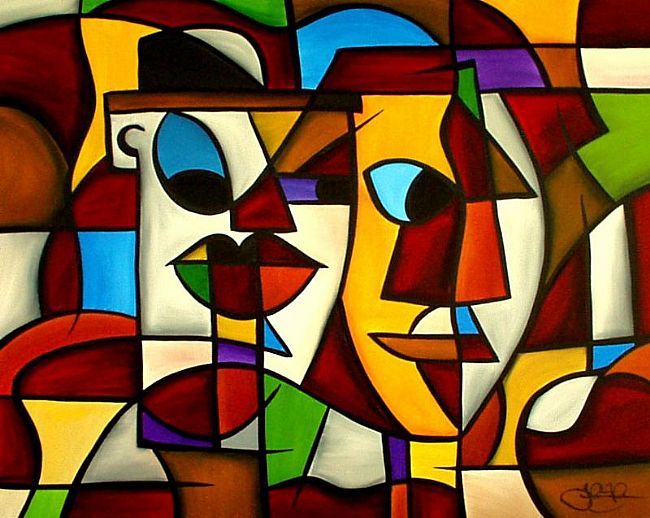}
			\includegraphics[width=0.49\textwidth]{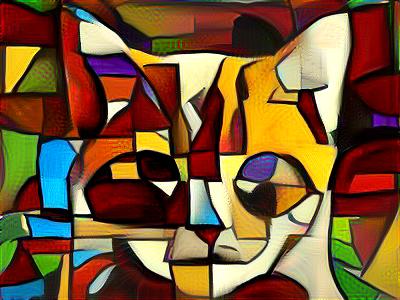}
			\includegraphics[width=0.49\textwidth]{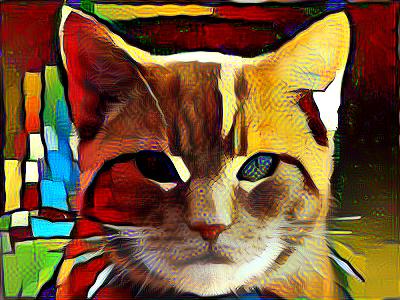}
			\includegraphics[width=0.49\textwidth]{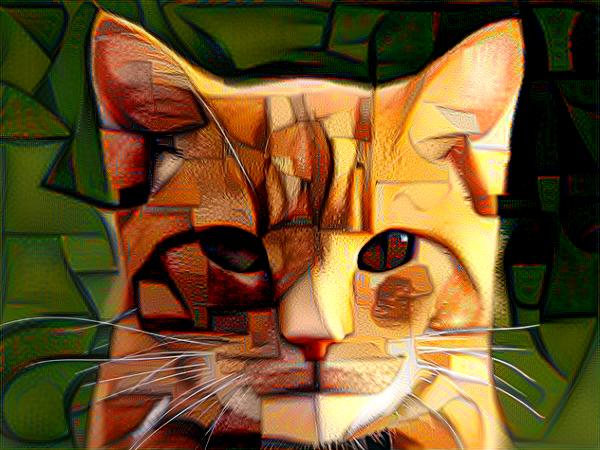}
		\end{center}
		\caption{Using $\{\texttt{conv1-1, conv2-1, conv4-1, conv4-2}\}$ as content and $\{\texttt{conv3-1, conv5-1}\}$ as style layers to repaint a photo of a cat. Style image (top-left, \cite{cubism}), full (top-right), low-weight (bottom-left) and partial style transfer (bottom-right). Notice that simply using a low style weight does not allow to reproduce the same result.}
		\label{fig:part}
	\end{figure*}

	For example, instead of using $\{\texttt{conv4-2}\}$ as the content layer and $\{\texttt{conv1-1,...,conv5-1}\}$ as the style layers, we could try to set $\{\texttt{conv1-1, conv2-1, conv4-1, conv4-2}\}$ as content and $\{\texttt{conv3-1, conv5-1}\}$ as style layers (see figure \ref{fig:part}). Notice how the partial style transfer managed to reshape the content and make it more rectangular (according to the style image) while preserving the original colors (which are mostly captured by the features in the bottom layers).
	
	We thus conclude that relaxing the bottom layer style constraints and enforcing them instead as the content constrains allows us to retain the colors and low-level features of the content photo and only transfer mid- to high-level properties of the style, combining them into a visually appealing result.

\section{Generating Styles}
\begin{figure*}
		\begin{center}
			\includegraphics[width=0.3\textwidth]{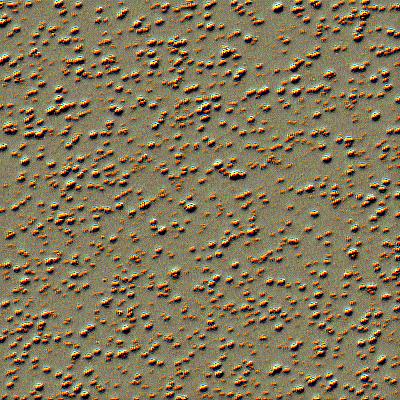}
			\includegraphics[width=0.3\textwidth]{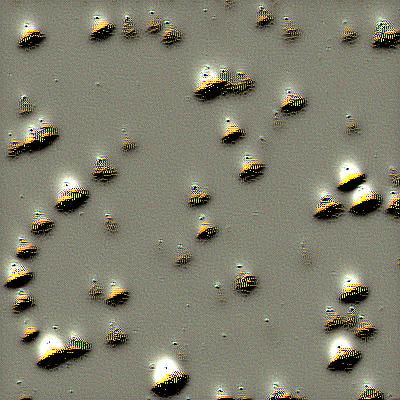}\\
			\includegraphics[width=0.3\textwidth]{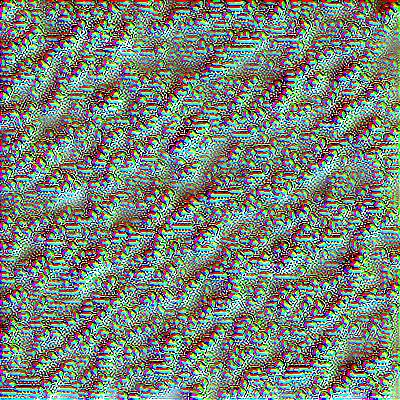}
			\includegraphics[width=0.3\textwidth]{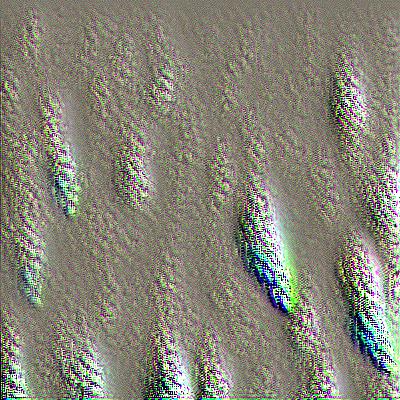}\\
			\includegraphics[width=0.3\textwidth]{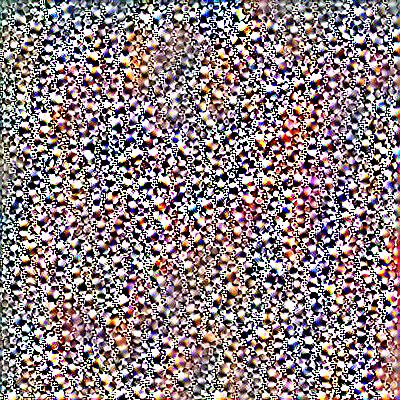}
			\includegraphics[width=0.3\textwidth]{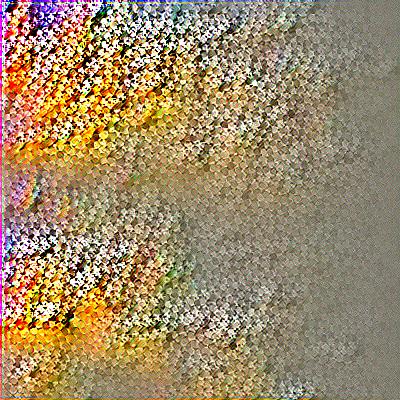}\\
			\includegraphics[width=0.3\textwidth]{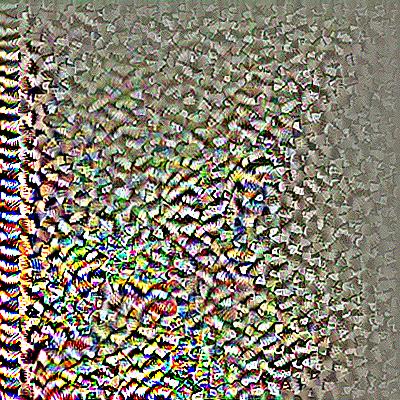}
			\includegraphics[width=0.3\textwidth]{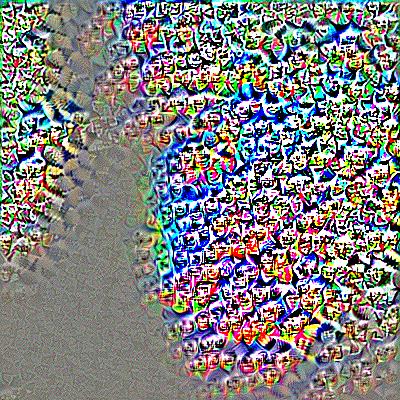}
		\end{center}
		\caption{Styles generated by targeting one Gram matrix $\mathcal{G}^l$ having a single non-zero entry. Top to bottom: layer $l$ from 1 to 4 of the target Gram matrix $\mathcal{G}^l$. Position of the non-zero element is either fixed (left) or generated randomly at each gradient descent iteration (right)}
		\label{fig:1hot}
	\end{figure*}

	In order to better understand the style space constructed in \cite{Gatys15} we draw inspiration from the Deep Visualization technique \cite{DD}. We disregard all the Gram matrices except for one with a single non-zero entry and descend from white noise without content constraints. The motivation is to understand what parts of style a single element can describe and verify if the Gram matrices present a basis in the style space qualitatively similar to the basis of VGG features in the space of natural images. By varying the non-zero element and its magnitude we can generate some curious textures with complexity increasing from the first to the fourth layer (see figure \ref{fig:1hot}).

\begin{figure*}
		\begin{center}
			\includegraphics[width=0.3\textwidth]{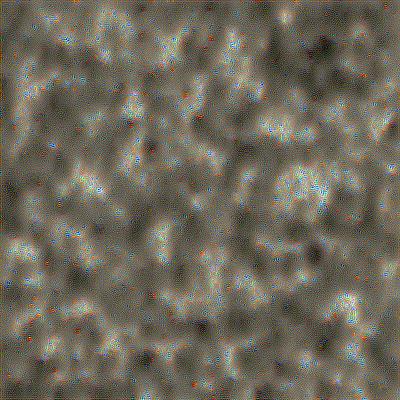}
			\includegraphics[width=0.3\textwidth]{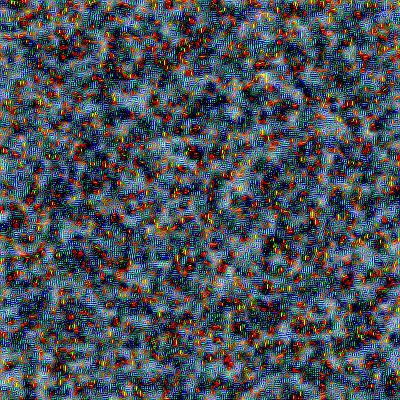}\\
			\includegraphics[width=0.3\textwidth]{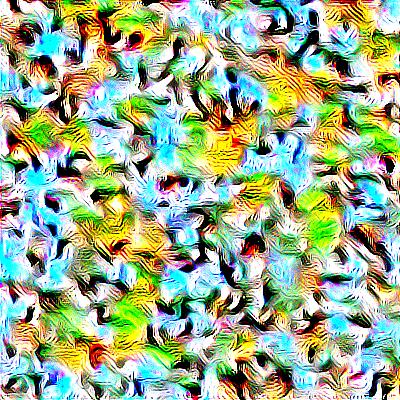}
			\includegraphics[width=0.3\textwidth]{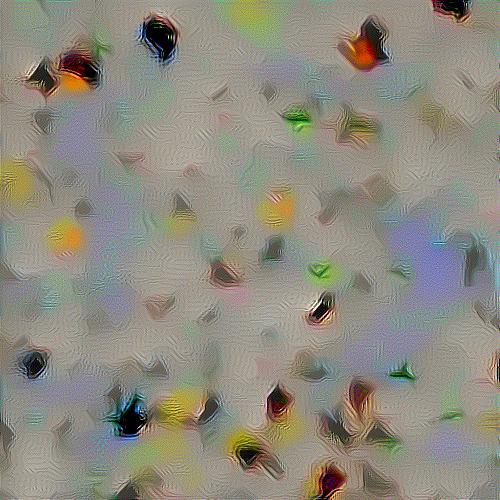}\\
			\includegraphics[width=0.3\textwidth]{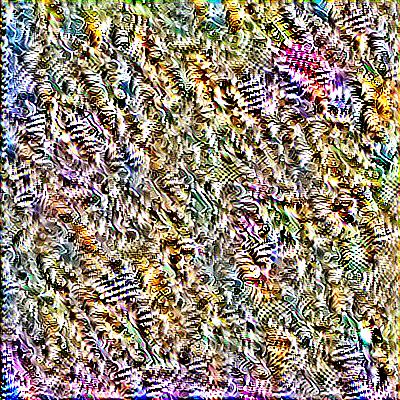}
			\includegraphics[width=0.3\textwidth]{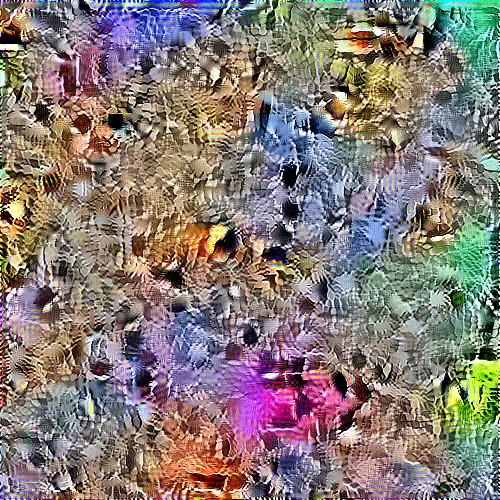}\\
			\includegraphics[width=0.3\textwidth]{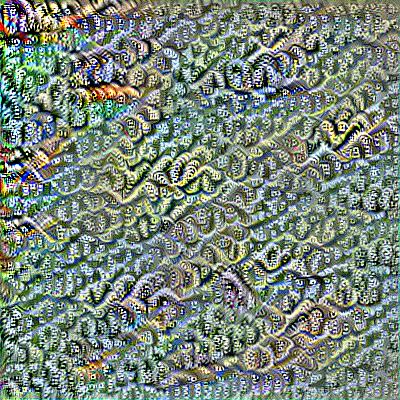}
			\includegraphics[width=0.3\textwidth]{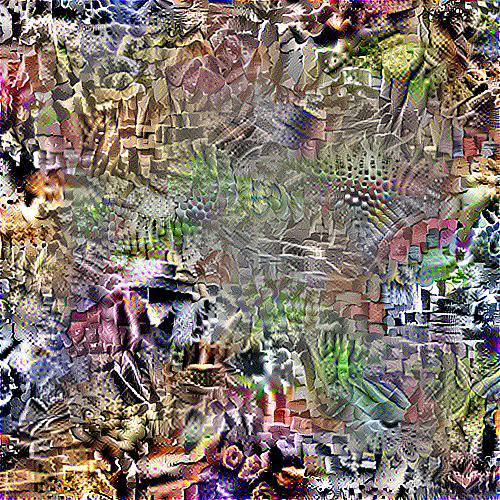}
		\end{center}
		\caption{Styles generated by targeting one random sparse Gram matrix $\mathcal{G}^l$. Top to bottom: layer $l$ from 1 to 4 of the target Gram matrix $\mathcal{G}^l$. The matrix is either fixed (left) or generated at each gradient descent iteration (right).}
		\label{fig:1layer}
	\end{figure*}

\begin{figure*}
		\begin{center}
			\includegraphics[width=0.49\textwidth]{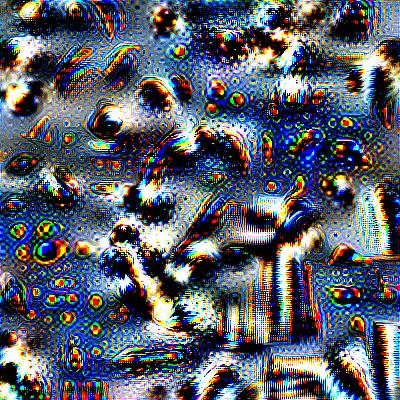}
			\includegraphics[width=0.49\textwidth]{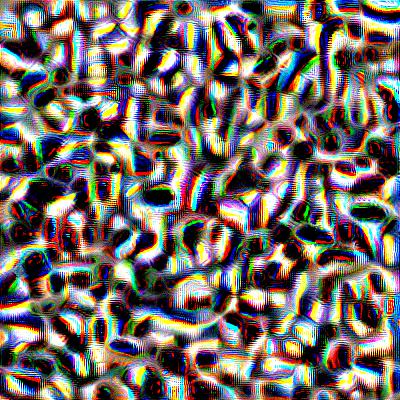}
			\includegraphics[width=0.49\textwidth]{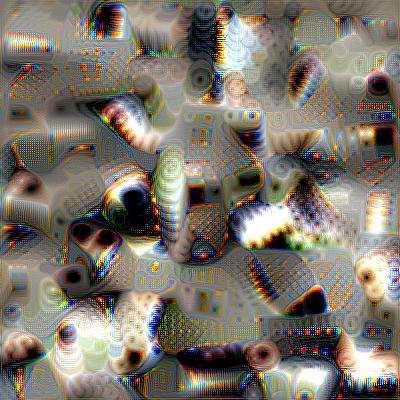}
			\includegraphics[width=0.49\textwidth]{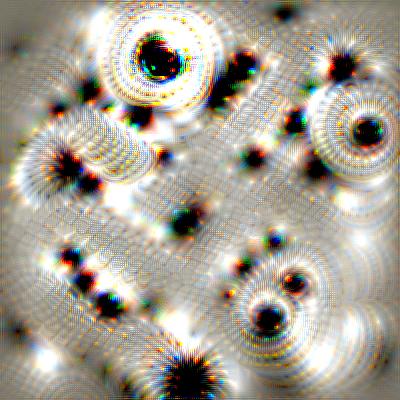}
		\end{center}
		\caption{Styles generated targeting multiple random (but fixed for the whole optimization procedure) sparse Gram matrices. Target matrices are at convolutional layers \{1, 2, 4\}, \{1, 2, 3\} (top row), \{1, 4\}, \{1, 2, 5\} (bottom row).}
		\label{fig:noisy}
	\end{figure*}

	Next we attempt to generate some more sophisticated styles. For this purpose we consider Gram matrices of a single painting and visualize its histogram. We observe that sparsity and amplitude of the Gram matrix elements increase from the first to the fifth layer (of course, this doesn't necessary need to generalize, yet it seems to be in accordance with the CNN paradigm where more complex and more discriminative features are constructed from simpler ones). Although it is impossible to judge about the distribution of Gram matrices describing styles with such a limited sample, we can try to mimic the sparsity and amplitude of Gram matrices representing the "Starry Night" by van Gogh \cite{gogh}. We generate a random matrix using absolute values of Gaussian distribution and apply a random sparse zero-one mask to it. We vary only two parameters: the variance of the Gaussian distribution and the sparsity of the mask. Interestingly enough, with such a simple construction, we were able to generate some intriguing textures (see figures \ref{fig:1layer} and \ref{fig:noisy})  suggesting that style density estimation and generation might be a promising research direction.

\section{Spatial Style Transfer}
	In \cite{Gatys15} each style layer $l$ with $k$ features introduces a $k\times k$ Gram matrix $\mathcal{G}^l$ with feature correlation entries $\mathcal{G}^l_{i j} = \langle F^l_i, F^l_j\rangle$. This makes the style completely invariant to the spatial configuration of the style image (which is by design).
	
	As an experiment, we suggest a new style representation designed to capture less of the artistic details and more of the spatial configuration of an image with roughly the same computational and storage complexity.
	
	We construct $\mathcal{G}^l$ matrices of size $k\times (2 X - 1) \times (2Y - 1)$ (where $X$ and $Y$ are the spatial dimensions of the layer $l$) with entries
	$$\mathcal{G}^l_{i}(x, y) = (F_i^l \ast F_i^l)(x, y),$$
	where $\ast$ stands for full 2D convolution. We thus impose a soft constraint on the feature distributions across the spatial coordinates (note that in principle we could store all pairwise convolutions of $F^l_i$ and $F^l_j$, but such an experiment would be computationally infeasible).
	
	In order for this objective to work we need to also rescale both content and style images to the same dimensions and modify the error derivative in \cite{Gatys15} accordingly:
	
	$$\frac{\partial E_l}{F^l_{i} (x, y)} = \frac{1}{N_l^2 M_l^2}\left[\left(\mathcal{G}^l - A^l\right)_i \circ F^l_i\right](x, y),$$
	
	where $\circ$ stands for valid correlation.
	
\begin{figure*}
		\begin{center}
			\includegraphics[width=0.4\textwidth]{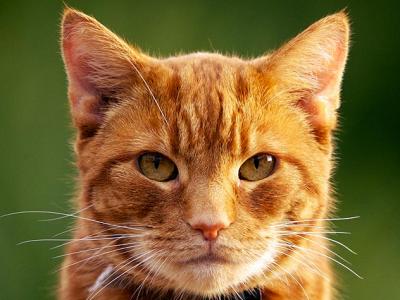}
			\includegraphics[width=0.4\textwidth]{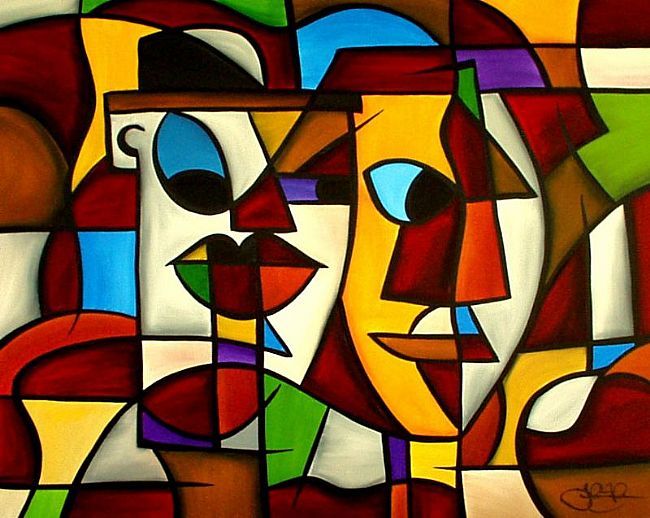}
			\includegraphics[width=0.4\textwidth]{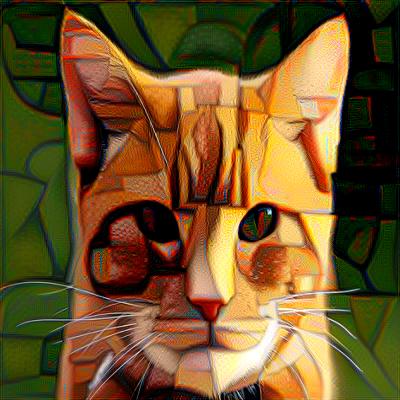}
			\includegraphics[width=0.4\textwidth]{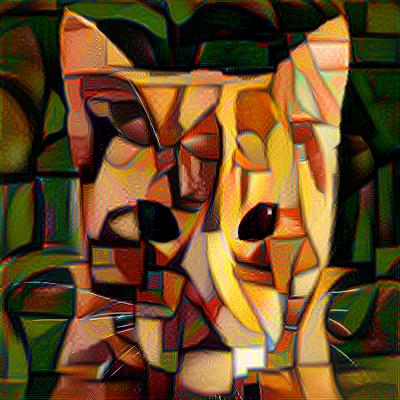}
			\includegraphics[width=0.4\textwidth]{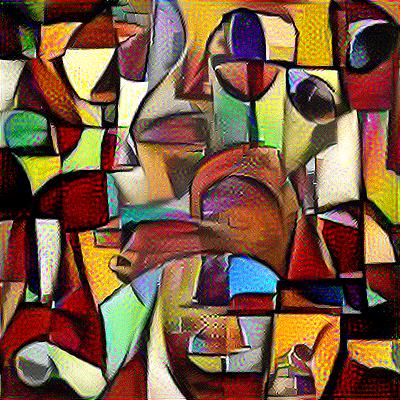}
			\includegraphics[width=0.4\textwidth]{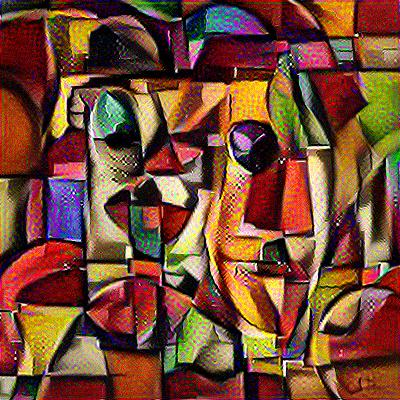}
		\end{center}
		\caption{Comparing the conventional and the ``spatial'' style transfer. Top row: content (left) and style (right); middle row: conventional (left) and ``spatial'' (right) style transfer; bottom row: style reconstruction from noise with no content constraint using conventional (left) and ``spatial'' (right) representations. Notice how the spatial style transfer gently imprints the scene structure into the photo. Reconstruction from noise demonstrates how it arrives at a rough but not exact configuration of the style scene.}
		\label{fig:corr}
	\end{figure*}

	We demonstrate this new approach on classic style transfer and on style reconstruction from noise (without content constraints) in figure \ref{fig:corr}.
	
	Notice how differently the proposed algorithm scales: it is easy to optimize on top layers (high $k$, small $X$ and $Y$) but is expensive on bottom layers (low $k$, huge $X$ and $Y$). This is contrary to the algorithm in \cite{Gatys15}, where the computation cost mostly depends on $k^2$ and thus  grows from bottom to top layers.

\section{Illumination and Season Transfer}\label{sec:ill}
	Not all artistic styles are transferred equally well by the algorithm in \cite{Gatys15}. If the style is highly repetitive and homogeneous over the whole style image (see abstract art, textures and patterns), it can be transferred with remarkable quality. However, once it becomes more subtle and varied within the image (see Renaissance, Baroque, Realism etc), style transfer falters. It fails to capture properties like the dramatic use of lighting, exaggerated face features etc. These properties get averaged-out, as the style representation is computed over the whole image.

	The same problem (i.e. global style representation) stands in the way of season and illumination transfer, as these properties change different elements of the scene differently. 
	
	We present some relatively successful examples of season and illumination transfer on images that are well aligned and are fairly repetitive in figures \ref{fig:tree}, \ref{fig:wood}, \ref{fig:mountain}, \ref{fig:city}.

\section{Towards Content-Aware Style Transfer}\label{sec:content}
\begin{figure*}
		\begin{center}
			\includegraphics[width=0.48\textwidth]{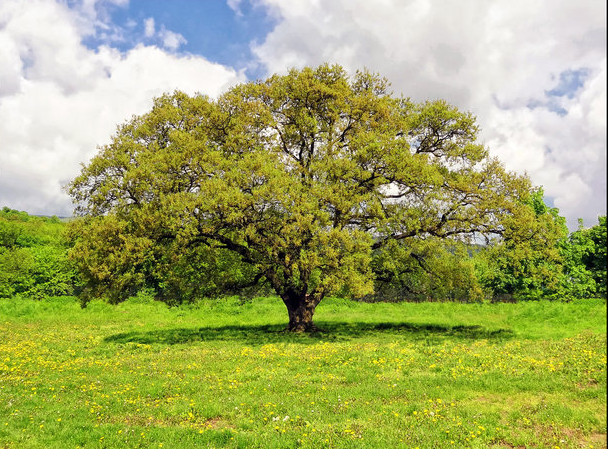}
			\includegraphics[width=0.5\textwidth]{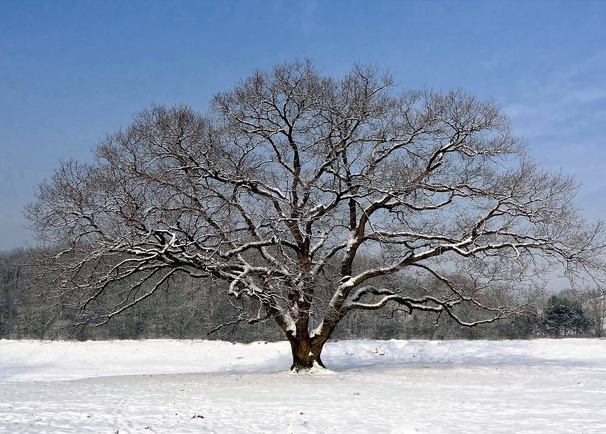}
			\includegraphics[width=0.33\textwidth]{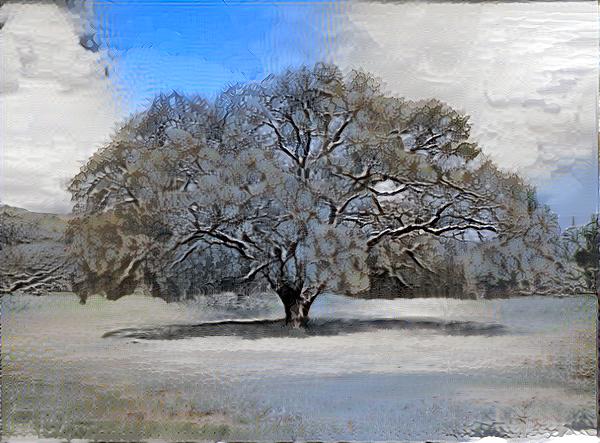}
			\includegraphics[width=0.33\textwidth]{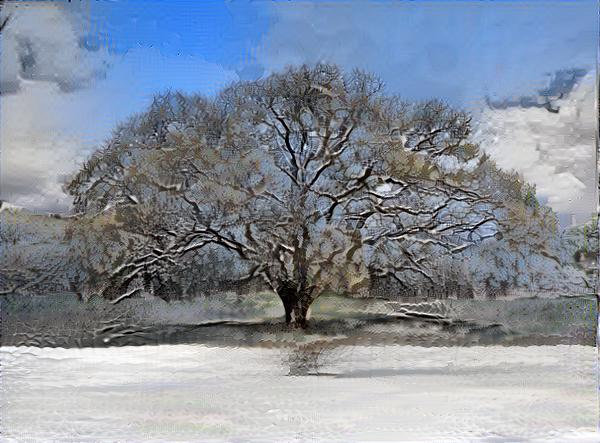}
			\includegraphics[width=0.33\textwidth]{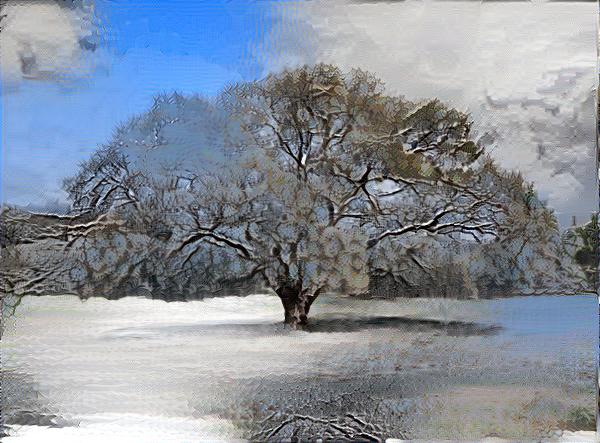}
		\end{center}
		\caption{An attempt of season transfer. Top row: content (left) and style (right). Bottom row: reconstruction from content (left), style (center) and noise (right). Notice how the problem of global style is especially well observed when descending from white noise, because random textures are generated everywhere and partially stay on the sky / ground, which is undesirable.}
		\label{fig:tree}
	\end{figure*}

\begin{figure*}
		\begin{center}
			\includegraphics[width=0.34\textwidth]{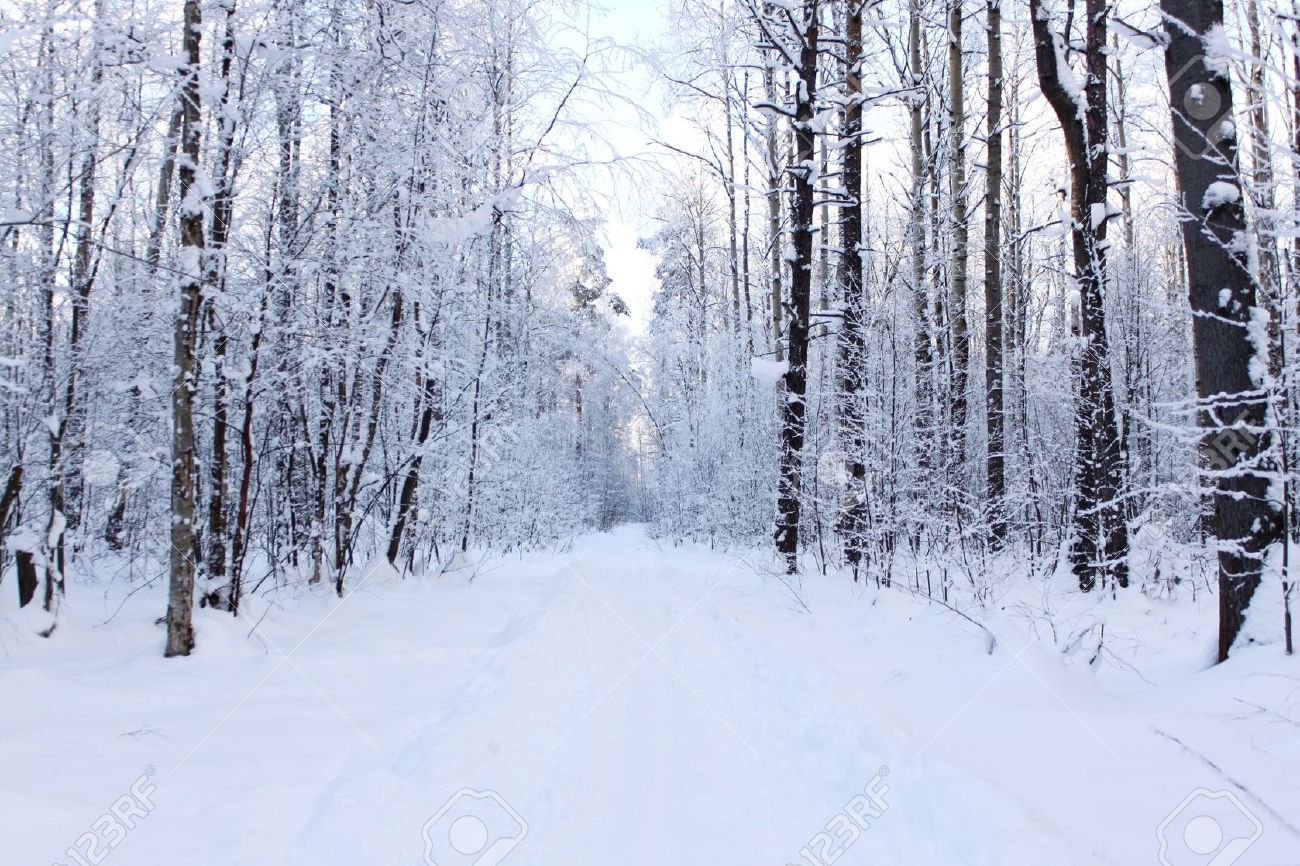}
			\includegraphics[width=0.303\textwidth]{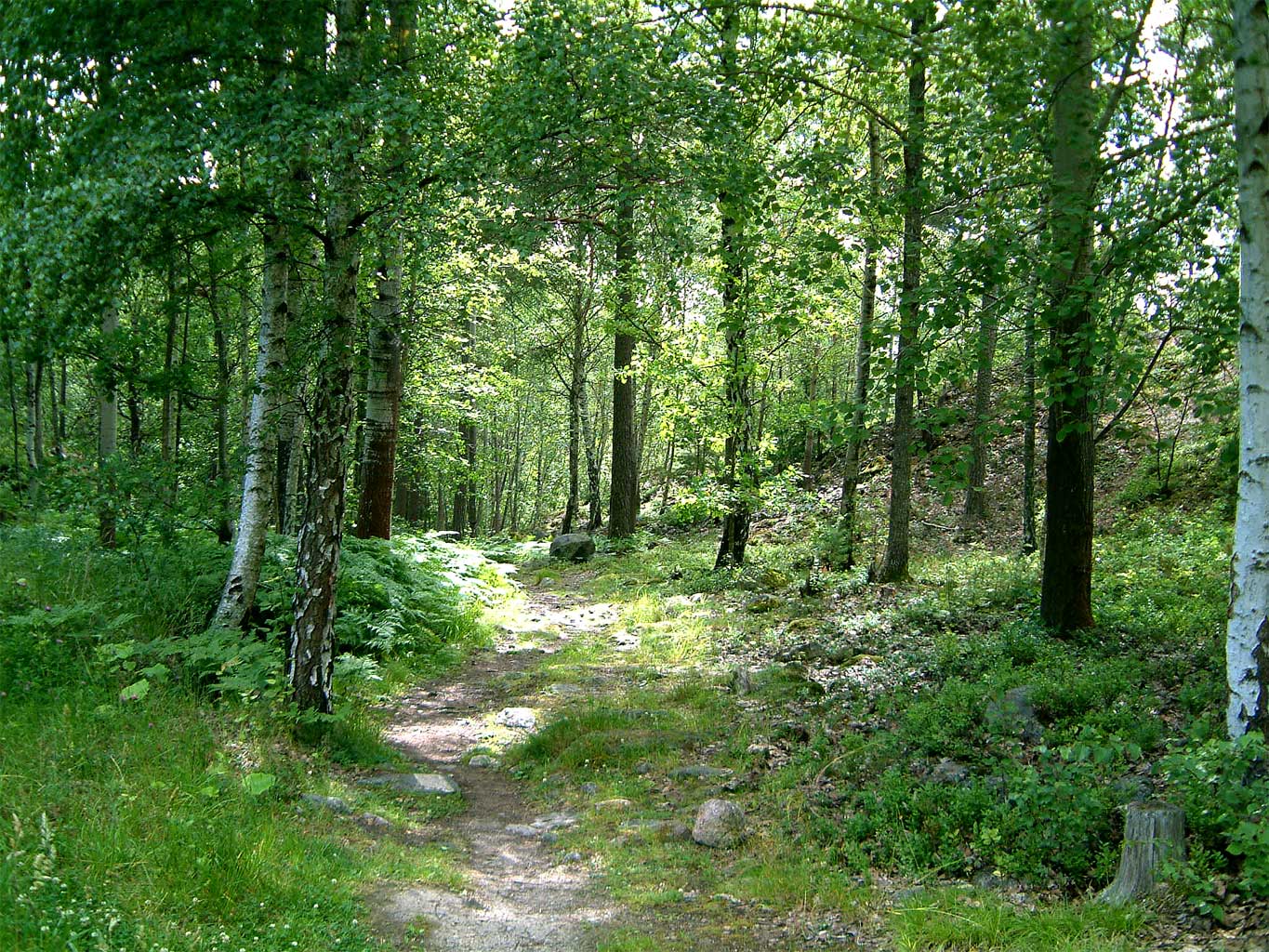}
			\includegraphics[width=0.34\textwidth]{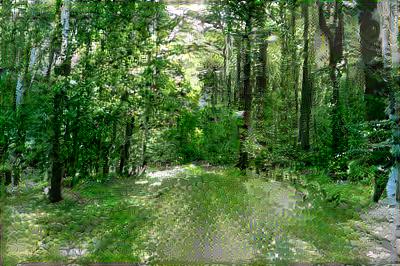}
		\end{center}
		\caption{Another attempt of season transfer. Content (left), style (center) and the transfer result (right).}
		\label{fig:wood}
	\end{figure*}

\begin{figure*}
		\begin{center}
			\includegraphics[width=0.27\textwidth]{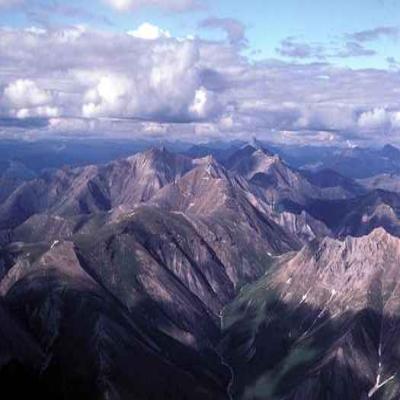}
			\includegraphics[width=0.43\textwidth]{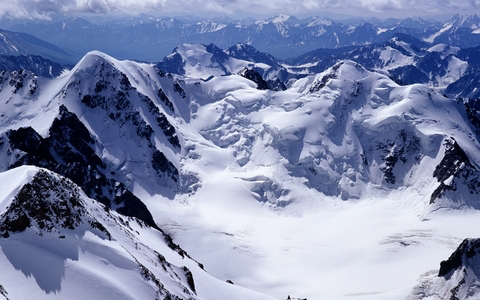}
			\includegraphics[width=0.27\textwidth]{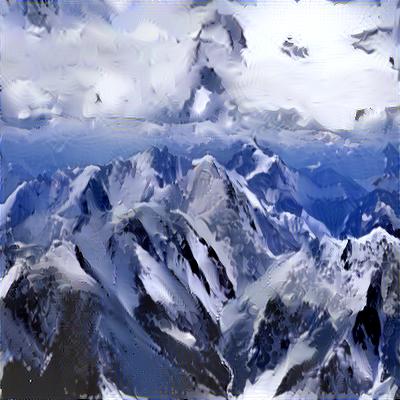}
		\end{center}
		\caption{Yet another attempt of season transfer. Content (left), style (center) and the transfer result (right).}
		\label{fig:mountain}
	\end{figure*}

\begin{figure*}
		\begin{center}
			\includegraphics[width=0.33\textwidth]{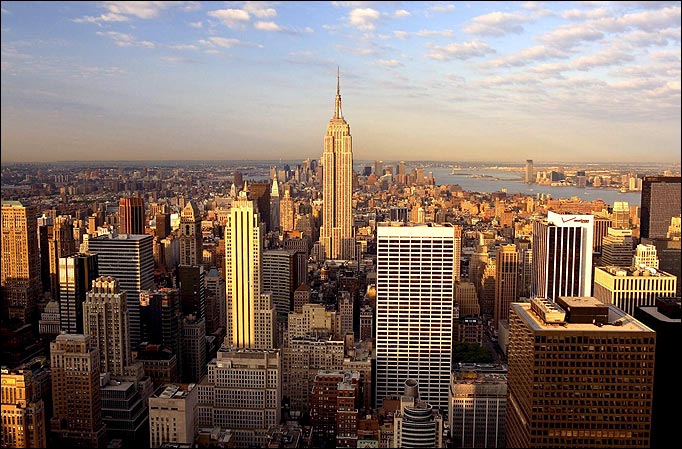}
			\includegraphics[width=0.33\textwidth]{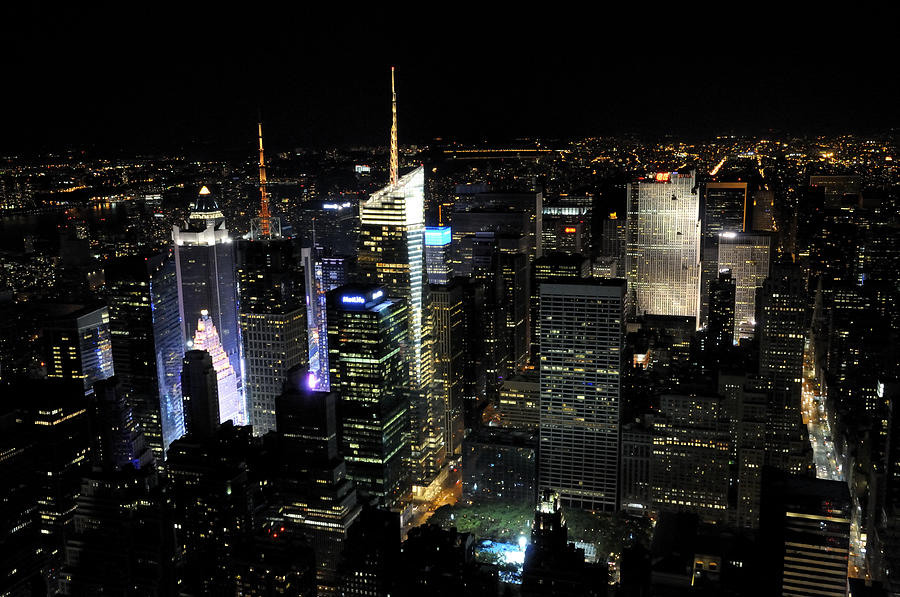}
			\includegraphics[width=0.33\textwidth]{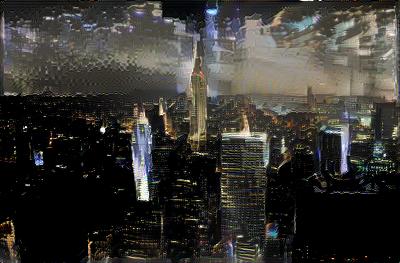}
		\end{center}
		\caption{An attempt of illumination transfer. Content (left), style (center) and the transfer result (right).}
		\label{fig:city}
	\end{figure*}

	The examples presented in figures \ref{fig:tree}, \ref{fig:wood}, \ref{fig:mountain}, \ref{fig:city} are quite bad, but they aren't \textit{too bad}. Indeed, one can easily spot a lot of regions where the texture was luckily transferred correctly. This serves to indicate that in principle the style representation developed in \cite{Gatys15} is capable of capturing photorealistic properties.

	What remains is developing a content-aware style transfer, i.e. transferring style according to the content matches between the image to be repainted and the style image. Below we discuss some possible leads towards implementing such a task.

	A direct approach could be to replace the global style values as defined in \cite{Gatys15}
	$$\mathcal{G}^l_{i j} = \langle F^l_i, F^l_j\rangle$$
	with localized values of
	$$\mathcal{G}_{i j}^l \left(\begin{matrix}
	           x \\
	           y 
	         \end{matrix}
	\right) = \sum_{-s^l \leqslant dx, dy \leqslant s^l} w\left(\begin{matrix}
	           dx \\
	           dy 
	         \end{matrix}
	\right) F^l_{i}\left(\begin{matrix}
	           x+dx \\
	           y+dy 
	         \end{matrix}
	\right)F^l_{j}\left(\begin{matrix}
	           x+dx \\
	           y+dy 
	         \end{matrix}
	\right)$$
	where we capture only feature correlations in a small ($s^l$) region around a point $(x, y)$ and we make the contributions decay as they get more distant from the point of interest:
	$$w \left(\begin{matrix}
	           dx \\
	           dy 
	         \end{matrix}
	\right) = \frac{1}{1 + dx^2 + dy^2}.$$

	We can then replace the global style loss
	$$E_l = \frac{1}{4 N_l^2 M_l^2} \left\|\mathcal{G}^l-A^l\right\|^2_2$$
	as defined in \cite{Gatys15} with a global style-content covariation loss
	
	$$E_l \sim \left\|\sum_{x, y}\left(\mathcal{F}^{c, l}_k\left(\begin{matrix}
	           x \\
	           y 
	         \end{matrix}
	\right) \mathcal{G}^l_{i j}\left(\begin{matrix}
	           x \\
	           y 
	         \end{matrix}
	\right) - \mathcal{P}^{c, l}_k \left(\begin{matrix}
	           x \\
	           y 
	         \end{matrix}
	\right)A_{i j}^l \left(\begin{matrix}
	           x \\
	           y 
	         \end{matrix}
	\right)\right)\right\|^2_2$$

	where $\mathcal{F}^{c, l}_k(x, y)$ is the weighted content response of neurons of filter $k$ in layer $c$ reachable from $(x, y,l)$ and the norm is that of a 3-dimensional tensor indexed with $i, j$ and $k$.

	Of course such an objective dramatically increases the computational cost of our problems and while could be efficiently parallelized, would probably still remain unfeasible. 
	
	In our implementation we make some very rough assumptions to test it: we consider $\mathcal{F}^c \equiv \mathcal{P}^c$ (content of images is aligned; we thus perform a locality-sensitive style transfer), $s = 0$ (pixel-wise style) and $w \equiv 1$ (no distance decay). This leads to a simplified expression of
	$$\mathcal{G}^l_{i j} = F^l_i \odot F^l_j$$
	and, consequently,
	$$\frac{\partial E_l}{\partial F^l_{i}(x, y)}=\frac{1}{N_l^2 M_l^2} \left[\left(\mathcal{G}^l_i - A^l_i\right)\left(\begin{matrix}x\\y\end{matrix}\right)\right]\left[{F^l_i}\left(\begin{matrix}x\\y\end{matrix}\right)\right]^T.$$
	
\begin{figure*}
		\begin{center}
			\includegraphics[width=0.13\textwidth]{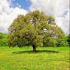}
			\includegraphics[width=0.13\textwidth]{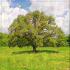}
			\includegraphics[width=0.13\textwidth]{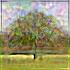}
			\includegraphics[width=0.13\textwidth]{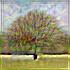}
			\includegraphics[width=0.13\textwidth]{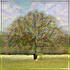}
			\includegraphics[width=0.13\textwidth]{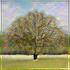}
			\includegraphics[width=0.13\textwidth]{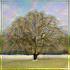}
			\includegraphics[width=0.13\textwidth]{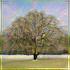}
			\includegraphics[width=0.13\textwidth]{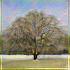}
			\includegraphics[width=0.13\textwidth]{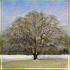}
			\includegraphics[width=0.13\textwidth]{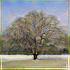}
			\includegraphics[width=0.13\textwidth]{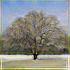}
			\includegraphics[width=0.13\textwidth]{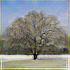}
			\includegraphics[width=0.13\textwidth]{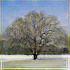}
			\includegraphics[width=0.13\textwidth]{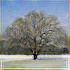}
			\includegraphics[width=0.13\textwidth]{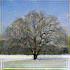}
			\includegraphics[width=0.13\textwidth]{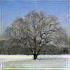}
			\includegraphics[width=0.13\textwidth]{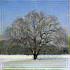}
			\includegraphics[width=0.13\textwidth]{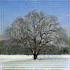}
			\includegraphics[width=0.13\textwidth]{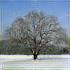}
			\includegraphics[width=0.13\textwidth]{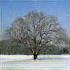}
		\end{center}
		\caption{Summer to winter transition using a very rough approximation of a locality-sensitive style transfer. Due to very rigid constraints the style gets basically painted over the content.}
		\label{fig:s2w}
	\end{figure*}

\begin{figure*}
		\begin{center}
			\includegraphics[width=0.13\textwidth]{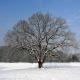}
			\includegraphics[width=0.13\textwidth]{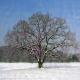}
			\includegraphics[width=0.13\textwidth]{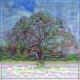}
			\includegraphics[width=0.13\textwidth]{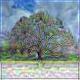}
			\includegraphics[width=0.13\textwidth]{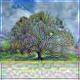}
			\includegraphics[width=0.13\textwidth]{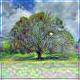}
			\includegraphics[width=0.13\textwidth]{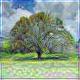}
			\includegraphics[width=0.13\textwidth]{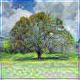}
			\includegraphics[width=0.13\textwidth]{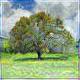}
			\includegraphics[width=0.13\textwidth]{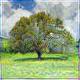}
			\includegraphics[width=0.13\textwidth]{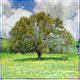}
			\includegraphics[width=0.13\textwidth]{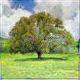}
			\includegraphics[width=0.13\textwidth]{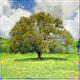}
			\includegraphics[width=0.13\textwidth]{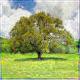}
			\includegraphics[width=0.13\textwidth]{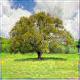}
			\includegraphics[width=0.13\textwidth]{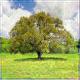}
			\includegraphics[width=0.13\textwidth]{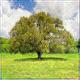}
			\includegraphics[width=0.13\textwidth]{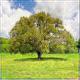}
			\includegraphics[width=0.13\textwidth]{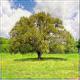}
			\includegraphics[width=0.13\textwidth]{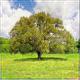}
			\includegraphics[width=0.13\textwidth]{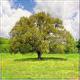}
		\end{center}
		\caption{Winter to summer transition using a very rough approximation of a locality-sensitive style transfer. As earlier, due to very rigid constraints the style gets basically painted over the content.}
		\label{fig:w2s}
	\end{figure*}

	We were only able to test this approach on small images (see figure \ref{fig:s2w} and \ref{fig:w2s}). Note that due to very rigid constraints the style picture basically gets painted over the content image.
	
	The next step within this approach would be to expand the style window $s^l$ and see whether a locality-sensitive style transfer is feasible. If it yields good results, the content-aware transfer could be investigated.
	
	We believe that if implemented well, such an algorithm could tackle more exquisite artistic styles, season transfer, illumination transfer, super-resolution and possibly many other applications.

\printbibliography

	\clearpage

\end{document}